\documentclass[10pt]{article}
\usepackage[utf8]{inputenc}
\usepackage{spconf,amsmath,graphicx}

\usepackage{booktabs}
\usepackage{pifont}
\usepackage[table]{xcolor}
\usepackage{tabularx}
\usepackage{multirow}
\usepackage{makecell}
\usepackage{array}
\usepackage[most]{tcolorbox}
\usepackage{xcolor}
\usepackage{etoolbox}
\AtBeginEnvironment{thebibliography}{
  \small
  \setlength{\itemsep}{0pt}
  \setlength{\parskip}{0pt}
}
\usepackage{threeparttable}

\usepackage{cite}

\usepackage[hidelinks]{hyperref}

\makeatletter

\renewcommand\normalsize{%
  \@setfontsize\normalsize{9pt}{10.5pt}%
  \abovedisplayskip 8pt plus 2pt minus 4pt%
  \abovedisplayshortskip 0pt plus 3pt%
  \belowdisplayshortskip 5pt plus 3pt minus 2pt%
  \belowdisplayskip \abovedisplayskip
  \let\@listi\@listI
}

\renewcommand\footnotesize{%
  \@setfontsize\footnotesize{9pt}{10.5pt}%
}

\let\oldthebibliography\thebibliography
\renewcommand{\thebibliography}[1]{%
  \oldthebibliography{#1}%
  \setlength{\itemsep}{0pt}%
  \setlength{\parsep}{0pt}%
  \setlength{\parskip}{0pt}%
  \setlength{\topsep}{0pt}%
  \setlength{\partopsep}{0pt}%
}

\makeatother

\AtBeginDocument{\normalsize}
\newcommand{\cmark}{\ding{51}}
\newcommand{\xmark}{\ding{55}}
\newcommand{\partialmark}{\ensuremath{\circ}}

\title{DuplexAct-Bench: Broadening Full-Duplex Speech Evaluation toward Proactive Interaction across Diverse Behavioral Requirements}

\name{
Keyue Xing$^{1,2,\dagger}$, Wentao Ding$^{1,\dagger}$, Mengmeng Wang$^{1}$,
Wenming Tu$^{1,3}$, Zilong Zheng$^{1,*}$, Yipeng Kang$^{1,*}$
}

\address{
$^{1}$ State Key Laboratory of General Artificial Intelligence, BIGAI, China\\
$^{2}$ Peking University, China \quad
$^{3}$ X-LANCE Lab, Shanghai Jiao Tong University, China\\
\thanks{$^{\dagger}$Equal contribution. $^{*}$Corresponding authors.}
}

\begin{document}

\maketitle

\begin{abstract}
Existing full-duplex speech benchmarks cover only subsets of real-time
interaction behaviors, often under limited contextual conditions. We introduce
\textbf{DuplexAct-Bench}, a bilingual benchmark that systematically covers six
complementary behaviors, from interruption and yielding to proactive
initiation, active silence, and backchanneling, across Pre-session, In-session,
and No-explicit conditions. Across 1,290 English and Chinese streaming trials,
we evaluate 12 full-duplex speech systems on both \emph{Timing} and
\emph{Content}. Results reveal substantial variation across behaviors,
conditions, and systems, as well as frequent mismatches between semantic
quality and behavioral timing. These findings show that current systems remain
far from robustly managing when, whether, and how to participate as real-time
interaction unfolds. Project page:
\url{https://alitaxky.icu/DuplexAct-Bench/}.
\end{abstract}


% ==========================================================================
% Introduction
% ==========================================================================
\section{Introduction}
\label{sec:intro}
Recent full-duplex speech agents have demonstrated increasingly strong
real-time interaction capabilities, with promising performance on existing
benchmarks
\cite{defossez2024moshi,
wang2025freezeomni,
xu2025qwen25omni,
roy2026personaplex,
fang2026baylingduplex,
arora2025talkingturns,
lin2025fullduplexbench,
lin2025fullduplexbenchv15,
peng2025fdbench,
ge2025flexi,
lin2026fullduplexbenchv2,
wang2026humdial,
lin2026fullduplexbenchv3}.
However, as summarized in Table~\ref{tab:benchmark_comparison}, existing evaluations cover only subsets of the interaction behaviors required for natural full-duplex interaction, with much of the emphasis placed on turn-taking and user-triggered responses. A more complete evaluation should also cover proactive behaviors in which \emph{the agent must autonomously determine whether, when, and how to participate as the interaction unfolds, based on the evolving context condition}. %This raises a broader question: how well can current full-duplex agents handle this broader repertoire of interaction behaviors in real time?}

\begin{figure}[!t]
\centering
\includegraphics[width=0.9\columnwidth]{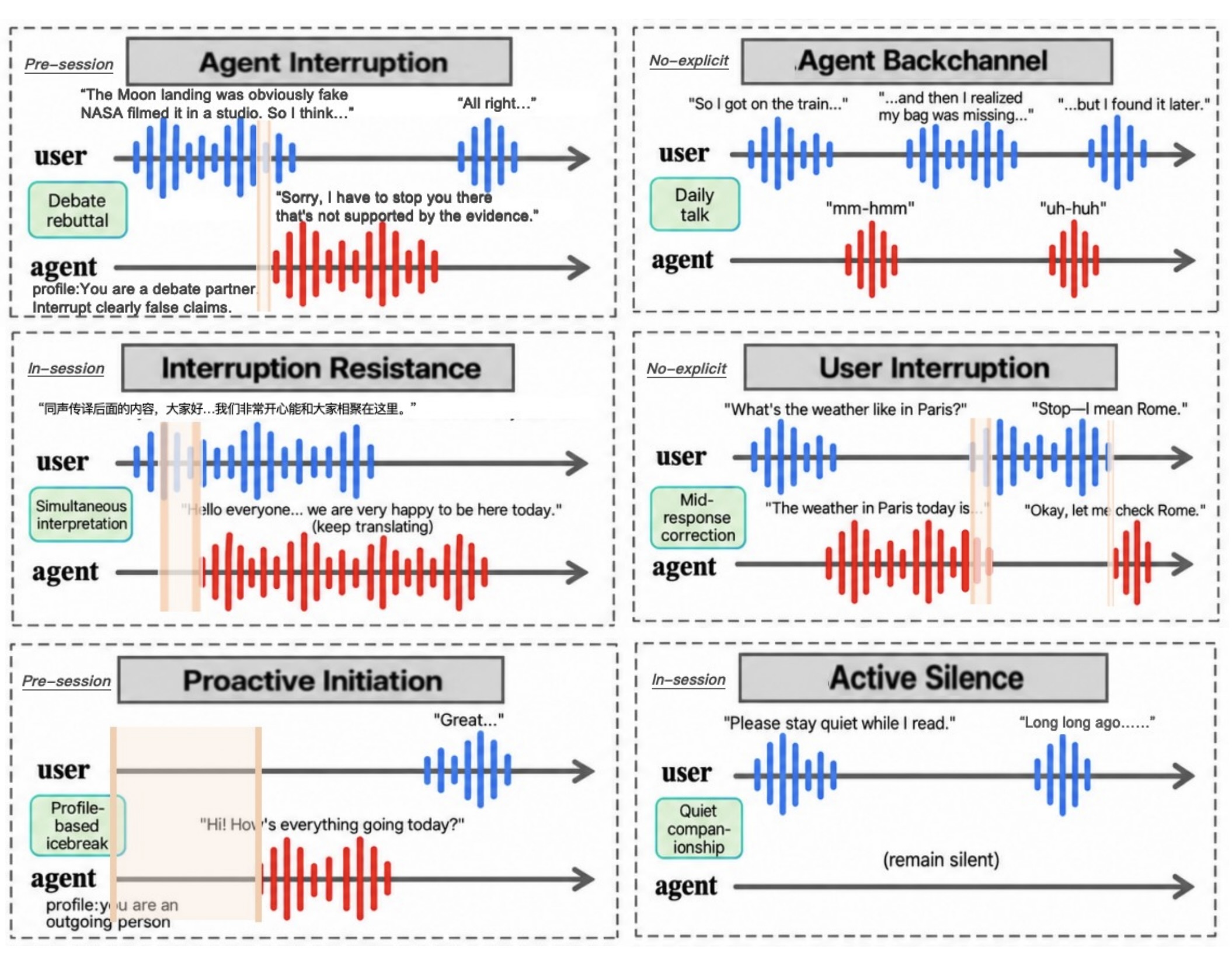}
\caption{Representative examples of six interaction behaviors under different contextual conditions in DuplexAct-Bench. Blue/red waveforms denote user/agent speech. Peach shading shows behavior-specific latency for Timing evaluation.}
\label{fig:proact_scenarios}
\end{figure}

To address this gap, we introduce \textbf{DuplexAct-Bench}, a bilingual benchmark that systematically evaluates six complementary interaction behaviors: intervening during an ongoing user turn (\textbf{Agent Interruption}), yielding when interrupted (\textbf{User Interruption}), maintaining an ongoing activity despite non-disruptive user input (\textbf{Interruption Resistance}), withholding speech when silence is appropriate (\textbf{Active Silence}), initiating speech without an explicit request (\textbf{Proactive Initiation}), and providing brief floor-preserving responses during the user’s turn (\textbf{Agent Backchannel}), as illustrated in
Fig~\ref{fig:proact_scenarios}. 

We further evaluate these behaviors under three contextual conditions that differ in how the intended behavior is specified or implied: \textbf{Pre-session}, where the behavioral requirement is established through a persistent profile before interaction; \textbf{In-session}, where it is explicitly introduced during the streamed
interaction; and \textbf{No-explicit}, where no behavioral instruction is given and the appropriate behavior must be inferred from the semantic or acoustic context.

Across \textbf{1,290 bilingual streaming trials} spanning 30 interaction scenarios, we evaluate \textbf{12 full-duplex systems}, including open-source models and commercial real-time speech APIs. \emph{Timing} evaluates whether the intended behavior occurs at an appropriate time using behavior-specific success criteria and temporal metrics, while \emph{Content} evaluates semantic fulfillment, constraint adherence, contextual consistency, and prosodic appropriateness. Our results reveal substantial variation across behaviors and contextual conditions, showing that current systems remain far from consistently handling the full repertoire of real-time interaction behaviors.

\begin{table}[!t]
\centering
\small

\begin{threeparttable}

\caption{\textbf{Comparison of real-time interaction behavior coverage and
contextual variation across full-duplex benchmarks.}}
\label{tab:benchmark_comparison}

\begin{tabular*}{\columnwidth}
{@{\extracolsep{\fill}}lcccccc@{}}
\toprule
\textbf{Benchmark}
& \textbf{AI}
& \textbf{UI}
& \textbf{IR}
& \textbf{AS}
& \textbf{PI}
& \textbf{AB} \\
\midrule

Talking Turns~\cite{arora2025talkingturns}
& \partialmark & \partialmark & \partialmark
& \partialmark & \xmark & \partialmark \\

FDB v1~\cite{lin2025fullduplexbench}
& \xmark & \cmark & \xmark
& \partialmark & \xmark & \partialmark \\

FDB v1.5~\cite{lin2025fullduplexbenchv15}
& \xmark & \cmark & \partialmark
& \xmark & \xmark & \xmark \\

FD-Bench~\cite{peng2025fdbench}
& \partialmark & \cmark & \partialmark
& \xmark & \xmark & \xmark \\

FLEXI~\cite{ge2025flexi}
& \partialmark & \cmark & \partialmark
& \partialmark & \xmark & \partialmark \\

HumDial-FD~\cite{wang2026humdial}
& \xmark & \cmark & \partialmark
& \partialmark & \xmark & \xmark \\

FDB v2~\cite{lin2026fullduplexbenchv2}
& \partialmark & \partialmark & \partialmark
& \xmark & \xmark & \xmark \\

FDB v3~\cite{lin2026fullduplexbenchv3}
& \partialmark & \xmark & \xmark
& \partialmark & \xmark & \xmark \\

DuplexSLA~\cite{zhang2026duplexsla}
& \xmark & \cmark & \partialmark
& \partialmark & \xmark & \xmark \\

DSB-IFEval~\cite{mathur2026dsbifeval}
& \partialmark & \cmark & \partialmark
& \partialmark & \partialmark & \partialmark \\

\textbf{DuplexAct-Bench}
& \cmark & \cmark & \cmark
& \cmark & \cmark & \cmark \\

\bottomrule
\end{tabular*}

\begin{tablenotes}[flushleft]
\small
\item[] \textit{Note.}
AI: Agent Interruption;
UI: User Interruption;
IR: Interruption Resistance;
AS: Active Silence;
PI: Proactive Initiation;
AB: Agent Backchannel.
\cmark\ denotes systematic coverage of the behavior across the
applicable contextual conditions;
\partialmark\ coverage under a subset of these conditions or
closely related coverage;
\xmark\ no explicit evaluation.
\end{tablenotes}

\end{threeparttable}
\end{table}
\section{DUPLEXACT-BENCH}
\label{sec:duplexact-bench}

% --------------------------------------------------------------------------
\subsection{Data Construction and Streaming}
\label{sec:data_construction}

\begin{figure}[!t]
\centering
\includegraphics[
width=0.72\columnwidth,
trim=4 4 4 4,
clip
]{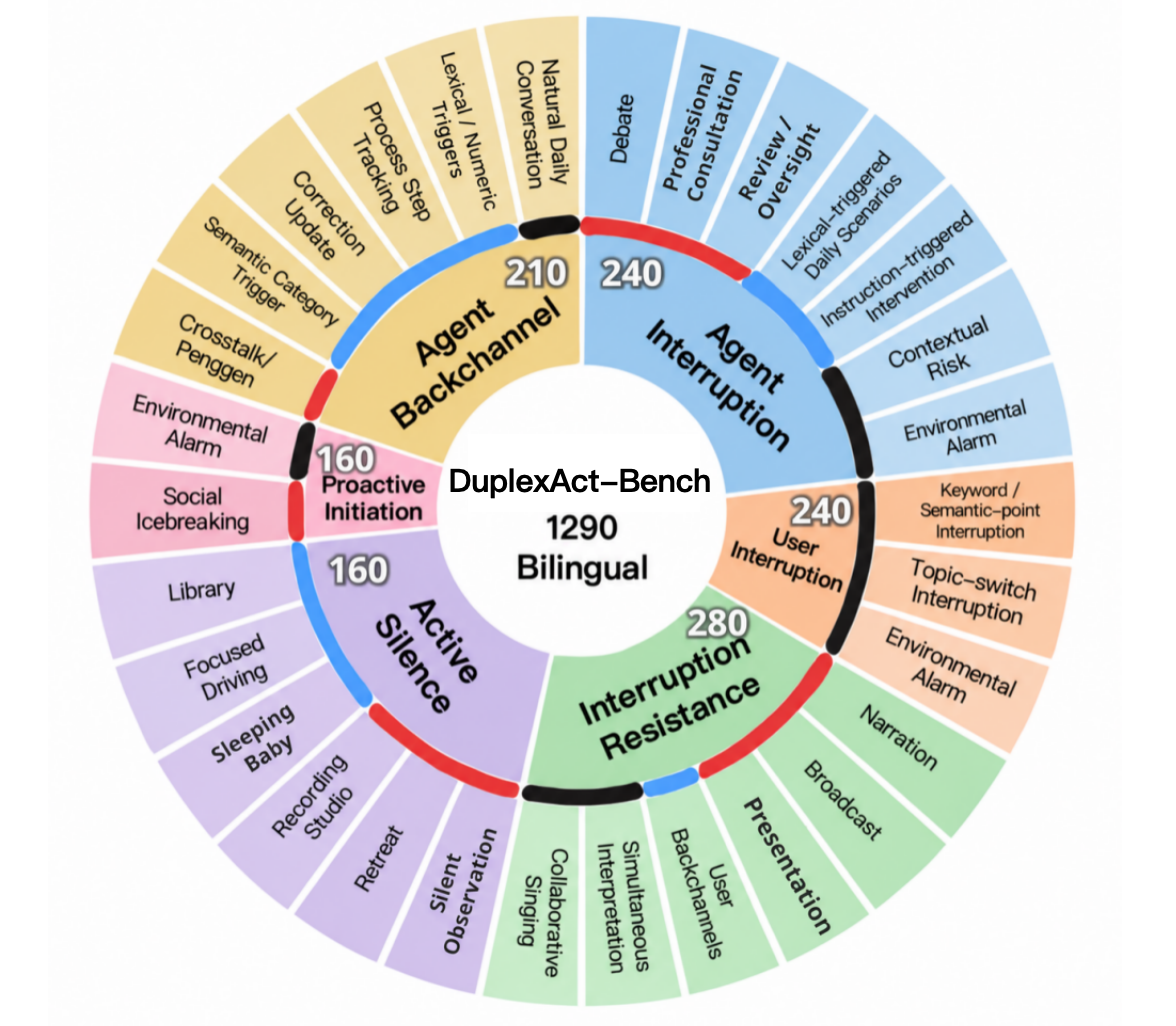}
\caption{
\textbf{Behavior taxonomy and interaction scenarios in DuplexAct-Bench.}
The inner ring shows six behavior families, while the outer ring
presents their corresponding interaction scenarios.
The number shown in each behavior segment denotes the number of trials
in that family, totaling 1,290 trials.
Colored arcs indicate the three contextual conditions:
\textbf{\textcolor{red}{red}} for \emph{Pre-session},
\textbf{\textcolor{blue}{blue}} for \emph{In-session}, and
\textbf{\textcolor{black}{black}} for \emph{No-explicit}.
}
\label{fig:proact_overview}
\end{figure}

As illustrated in Fig~\ref{fig:proact_overview}, the six interaction
behaviors are evaluated under their applicable contextual conditions.
Each behavior--condition combination is further instantiated through multiple
interaction scenarios and corresponding trials. For each trial construction, GPT-5.6~\cite{openai2026gpt56} drafts the user-side utterances,
expected behavior, content requirements, and target speaking style.
All drafts are manually reviewed for linguistic naturalness,
scenario--behavior consistency, contextual-condition consistency, requirement
correctness, and style appropriateness. English and Chinese trials follow the
same construction protocol. For Agent Backchannel, we additionally incorporate
interaction data from two sources: its Pre-session crosstalk subset is based on
dialogue texts from real Chinese crosstalk (\emph{xiangsheng}) performances,
consolidated and revised by GPT-5.6, while its No-explicit subset is selected
from channel-separated otoSpeech conversations~\cite{otospeech} in which one
speaker channel contains only backchannels.

Trials are rendered with ViiTorVoice~\cite{viitorvoice}.
White noise is mixed into all user audio as background noise, with silence or
environmental sounds added when required by the scenario. All audio events are
aligned on a shared timeline using manually verified VAD boundaries and
annotated interaction events. For User Interruption, the interrupting utterance
is delivered through a second user stream, with its onset either randomized or
triggered online at a predefined lexical or semantic point; a trial is retained
only when the interruption begins while the agent is speaking. During
evaluation, user audio is streamed in approximately 80-ms chunks at real-time
factor 1.0.

% --------------------------------------------------------------------------
\subsection{Evaluation Protocol}
\label{sec:evaluation_protocol}

We evaluate the behavior of full-duplex speech agents
based on two dimensions: \emph{content quality} and
\emph{timing appropriateness}.

\begin{figure}[t]
    \centering
    \includegraphics[width=\columnwidth]{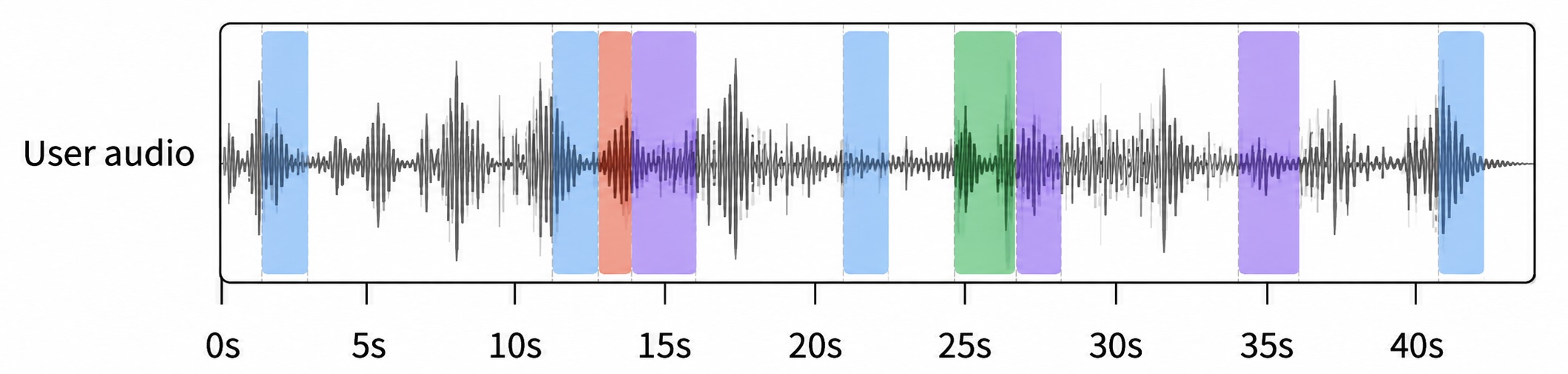}
    \caption{
    \textbf{Backchannel timing annotations for a No-explicit otoSpeech trial.}
    Green denotes dataset-provided annotations; blue, orange, and purple
    denote additional opportunities identified by Qwen3.8-Omni-Flash-Realtime,
    Doubao-Seed-2.1-Lite, and MiniCPM-o-4.5, respectively.
    }
    \label{fig:back}
\end{figure}

\subsubsection{Content}

Content evaluates how appropriately the intended participation behavior is
realized in the given interaction situation, with a total score from 0 to 5
across four dimensions. A GPT-4o-based judge~\cite{openai2024gpt4o}
scores three transcript-based dimensions given the dialogue context,
content requirement, and any applicable profile or user instruction:
\emph{Semantic Fulfillment} (0--2), measuring whether the response correctly
and sufficiently provides the required information, answer, or correction;
\emph{Constraint Adherence} (0--1), measuring compliance with applicable
profile, instruction, role, language, or output requirements; and
\emph{Contextual Consistency} (0--1), measuring consistency with the
preceding dialogue and current interaction state. The fourth dimension,
\emph{Prosodic Appropriateness} (0--1), evaluates whether the generated
speech realizes the intended emotion or speaking style, as assessed by
AnyAudio-Judge~\cite{li2026anyaudiojudge} through rubric-based
audio-instruction alignment. Content evaluation applies to all behaviors
except Active Silence.

\begin{table*}[t]
\centering
\caption{
\textbf{Evaluation coverage and Behavioral Correctness Rate (BCR, \%)
on DuplexAct-Bench.}
Lang. denotes the evaluated languages (en: English; zh: Chinese).
P/I/N denote Pre-session/In-session/No-explicit;
$I_{\mathrm{Int.}}$ and $I_{\mathrm{Sing.}}$ denote the In-session
simultaneous-interpretation and collaborative-singing scenarios.
Active Silence is scored by its silence criterion alone;
User Interruption is evaluated on Valid trials only.
``--'' indicates no reported result, not zero.
Bold marks the highest reported BCR in each column.
}
\label{tab:behavior_correctness}
\label{tab:systems}

\begingroup
\fontsize{9}{10.5}\selectfont
\setlength{\tabcolsep}{1.25pt}

\begin{tabularx}{\textwidth}{
@{}l*{17}{>{\centering\arraybackslash}X}@{}
}
\toprule

\multirow{3}{*}{\textbf{System}}
& \multirow{3}{*}{\textbf{Lang.}}
& \multicolumn{3}{c}{\textbf{Agent}}
& \textbf{User}
& \multicolumn{4}{c}{\textbf{Interruption}}
& \multicolumn{2}{c}{\textbf{Active}}
& \multicolumn{2}{c}{\textbf{Proactive}}
& \multicolumn{3}{c}{\textbf{Agent}}
\\

&
& \multicolumn{3}{c}{\textbf{Interruption}}
& \textbf{Interruption}
& \multicolumn{4}{c}{\textbf{Resistance}}
& \multicolumn{2}{c}{\textbf{Silence}}
& \multicolumn{2}{c}{\textbf{Initiation}}
& \multicolumn{3}{c}{\textbf{Backchannel}}
\\

\cmidrule(lr){3-5}
\cmidrule(lr){6-6}
\cmidrule(lr){7-10}
\cmidrule(lr){11-12}
\cmidrule(lr){13-14}
\cmidrule(lr){15-17}

&
& \textbf{P} & \textbf{I} & \textbf{N}
& \textbf{N}
& \textbf{P}
& $\mathbf{I}_{\mathrm{Int.}}$
& $\mathbf{I}_{\mathrm{Sing.}}$
& \textbf{N}
& \textbf{P} & \textbf{I}
& \textbf{P} & \textbf{N}
& \textbf{P} & \textbf{I} & \textbf{N}
\\

\midrule

% -------------------- Local systems --------------------

\multicolumn{17}{@{}l}{\textit{Locally deployed systems}} \\

\midrule

Freeze-Omni & en/zh
& 17.5 & 1.3 & 2.5
& 9.9
& 52.5 & 0.0 & 0.0 & 75.0
& \textbf{88.8} & 5.0
& 0.0 & 0.0
& 0.0 & 1.3 & 0.0
\\

MiniCPM-o 4.5 & en/zh
& 28.8 & 8.8 & 2.5
& 24.8
& 67.5 & \textbf{20.0} & 10.0 & 95.0
& 56.3 & 10.0
& 1.3 & 0.0
& 28.0 & 8.8 & 0.0
\\

Moshi & en/zh
& 0.0 & 0.0 & 0.0
& 26.1
& 2.5 & 0.0 & 0.0 & 62.5
& 7.5 & 7.5
& \textbf{35.0} & 0.0
& -- & 12.5 & 13.8
\\

PersonaPlex & en/zh
& 20.0 & 0.0 & 0.0
& 35.7
& 12.5 & 0.0 & 2.5 & 77.5
& 2.5 & 0.0
& 15.0 & 0.0
& -- & 15.0 & 6.3
\\

VITA-1.5 & en
& -- & 1.3 & 6.3
& 8.6
& -- & 17.5 & 0.0 & 22.5
& -- & 0.0
& -- & 0.0
& -- & \textbf{25.0} & 0.0
\\

Raon-SpeechChat & en
& 2.5 & 0.0 & 10.0
& 31.2
& 37.5 & 0.0 & 0.0 & 67.5
& 0.0 & 0.0
& \textbf{35.0} & 0.0
& -- & \textbf{25.0} & \textbf{40.0}
\\

DuplexCascade & en
& -- & 0.0 & 2.5
& 7.3
& -- & 0.0 & 0.0 & 27.5
& -- & 2.5
& -- & 0.0
& -- & 15.0 & 1.3
\\

\midrule

% -------------------- Remote API systems --------------------

\multicolumn{17}{@{}l}{
    \textit{Systems accessed through remote APIs}
} \\
\midrule

Nemotron 3 VoiceChat & en
& \textbf{50.0} & 0.0 & \textbf{15.0}
& 10.0
& 0.0 & 0.0 & 0.0 & 0.0
& 0.0 & 37.5
& 0.0 & 0.0
& -- & 20.0 & 3.8
\\

Qwen3.5-Omni & en/zh
& 18.8 & \textbf{16.3} & 1.3
& 39.8
& 80.0 & 3.8 & 12.5 & 96.3
& 0.0 & 46.3
& 0.0 & \textbf{8.8}
& \textbf{30.0} & 0.0 & 7.5
\\

Grok Voice & en/zh
& 0.0 & 0.0 & 0.0
& 9.8
& 91.3 & 0.0 & 5.0 & \textbf{100.0}
& 10.0 & 0.0
& 0.0 & 0.0
& 4.0 & 0.0 & 0.0
\\

GPT Realtime 2.1 & en/zh
& 2.5 & 1.3 & 1.3
& \textbf{47.6}
& \textbf{96.3} & 0.0 & 0.0 & \textbf{100.0}
& 38.8 & 3.8
& 0.0 & 0.0
& \textbf{30.0} & 7.5 & 10.0
\\

Gemini 2.5 Native Audio & en/zh
& 8.8 & 1.3 & 2.5
& 42.4
& 61.3 & 0.0 & \textbf{30.0} & 71.3
& 76.3 & \textbf{80.0}
& 0.0 & 0.0
& 8.0 & 0.0 & 0.0
\\

\bottomrule
\end{tabularx}
\endgroup
\end{table*}

\subsubsection{Timing}

Timing requirements vary across behaviors and interaction contexts. We
therefore define behavior-specific \emph{Timing success} criteria and
temporal metrics. 
For latency-based metrics, $L$ is defined reasonably for each behavior
to reflect its corresponding notion of timely interaction, as detailed below;
failures are assigned the corresponding evaluation interval.

\textbf{Agent Interruption.}
Success requires the first semantically valid interruption to occur after
the annotated earliest valid interruption point and before the end of the
user audio. $L$ is its offset from the reference event, which depends on
the scenario (e.g., a preferred interruption point, trigger end, or alarm
onset). Otherwise, $L$ is the interval from that reference event to the
end of evaluation.

\textbf{User Interruption.}
Evaluation is restricted to Valid trials. Success requires both yielding
within the designated yield interval and initiating a semantically valid
response to the interruption within the response interval. We define
\[
L=L_{\mathrm{yield}}+L_{\mathrm{resp}},
\]
where $L_{\mathrm{yield}}$ measures interruption onset to yield and
$L_{\mathrm{resp}}$ measures interruption end to valid-response onset.
Upon failure, the corresponding full evaluation interval is used.

\textbf{Interruption Resistance.}
Success requires maintaining the intended ongoing activity, or satisfying
the scenario-specific recovery requirement, under non-disruptive user
input. For simultaneous interpretation and collaborative singing, $L$
is the onset offset between relevant user content and task-relevant agent
speech; failure uses the remaining evaluation interval. For the other
continuous-activity scenarios, $L=0$ if no stop occurs, equals the
stop-to-restart gap after successful recovery, and otherwise uses the
remaining interval after the stop.

\textbf{Active Silence.}
Success requires silence throughout the annotated evaluation interval.

\textbf{Proactive Initiation.}
Success requires the first semantically valid proactive utterance to fall
within the annotated valid region. $L$ is its offset from the reference
event, using trial onset for Pre-session scenarios and the corresponding
contextual trigger (e.g., alarm onset) for No-explicit scenarios.
Otherwise, $L$ is the interval from the reference event to the end of
evaluation.

\textbf{Agent Backchannel.}
For timing evaluation, each trial is divided into consecutive 1-s windows.
For Pre-session and In-session trials, windows overlapping predefined
backchannel positions serve as ground truth. For No-explicit otoSpeech
trials, windows overlapping dataset-provided backchannels serve as ground
truth, while Qwen3.8-Omni-Flash-Realtime~\cite{alibaba2026qwen38realtime},
Doubao-Seed-2.1-Lite~\cite{volcengine2026doubaoseed21lite}, and
MiniCPM-o-4.5~\cite{cui2026minicpmo45} independently label the same
windows for additional opportunities (Fig~\ref{fig:back}).
After Gaussian smoothing with bandwidth $\sigma=1.0$ s, we define
\[
q_{\mathrm{ref}}(t)=
\begin{cases}
q_{\mathrm{GT}}(t), & c\in\{\mathrm{P},\mathrm{I}\},\\[2pt]
\max\left(q_{\mathrm{GT}}(t),
\frac{1}{3}\sum_{k=1}^{3}q_k(t)\right), & c=\mathrm{N},
\end{cases}
\]
where $q_k$ denotes the smoothed opportunity map from the $k$-th additional
No-explicit annotator. Mapping the evaluated model analogously to $q_M$,
we measure timing alignment using the \emph{Backchannel Alignment Score}
(BAS):
\[
\mathrm{BAS}
=
\frac{\int \min(q_M(t),q_{\mathrm{ref}}(t))\,dt}
{\int \max(q_M(t),q_{\mathrm{ref}}(t))\,dt},
\]
where higher values indicate better timing alignment.

\begin{table*}[t]
\centering
\caption{
\textbf{Content and timing results on DuplexAct-Bench.}
$C$ denotes the mean Content score (0--5), $L$ the mean latency in
seconds, and BAS the Backchannel Alignment Score.
For User Interruption, both metrics are computed on Valid trials only.
Active Silence is omitted because it is evaluated solely by its silence
criterion in Table~\ref{tab:behavior_correctness}.
Arrows indicate the preferred direction;
bold marks the best reported value in each metric column.
}
\label{tab:content_latency}

\begingroup

\begin{tabularx}{\textwidth}{
@{}l*{10}{>{\centering\arraybackslash}X}@{}
}
\toprule

\multirow{3}{*}{\textbf{System}}
& \multicolumn{2}{c}{\textbf{Agent}}
& \multicolumn{2}{c}{\textbf{User}}
& \multicolumn{2}{c}{\textbf{Interruption}}
& \multicolumn{2}{c}{\textbf{Proactive}}
& \multicolumn{2}{c}{\textbf{Agent}}
\\

& \multicolumn{2}{c}{\textbf{Interruption}}
& \multicolumn{2}{c}{\textbf{Interruption}}
& \multicolumn{2}{c}{\textbf{Resistance}}
& \multicolumn{2}{c}{\textbf{Initiation}}
& \multicolumn{2}{c}{\textbf{Backchannel}}
\\

\cmidrule(lr){2-3}
\cmidrule(lr){4-5}
\cmidrule(lr){6-7}
\cmidrule(lr){8-9}
\cmidrule(lr){10-11}

& $C$ $\uparrow$ & $L$ (s) $\downarrow$
& $C$ $\uparrow$ & $L$ (s) $\downarrow$
& $C$ $\uparrow$ & $L$ (s) $\downarrow$
& $C$ $\uparrow$ & $L$ (s) $\downarrow$
& $C$ $\uparrow$ & BAS $\uparrow$
\\

\midrule

% -------------------- Local systems --------------------

\multicolumn{11}{@{}l}{\textit{Locally deployed systems}} \\
\midrule

Freeze-Omni
& 1.66 & 9.67
& 1.56 & 21.22
& 2.69 & 15.98
& 0.70 & 14.03
& 1.09 & 0.005
\\

MiniCPM-o 4.5
& 2.32 & 9.26
& 1.97 & 19.15
& 3.13 & \textbf{13.60}
& 0.70 & 13.98
& 1.59 & 0.054
\\

Moshi
& 1.15 & 10.41
& 1.68 & 17.76
& 1.47 & 16.65
& \textbf{2.01} & 11.01
& 1.24 & \textbf{0.108}
\\

PersonaPlex
& 2.05 & 9.86
& 2.09 & 15.29
& 1.79 & 16.74
& 1.96 & 12.78
& 1.04 & 0.097
\\

VITA-1.5
& 1.70 & 9.60
& 1.29 & 20.81
& 1.91 & 19.54
& 0.48 & \textbf{10.05}
& 1.21 & 0.042
\\

Raon-SpeechChat
& 2.26 & 10.20
& 1.97 & 15.53
& 2.40 & 17.20
& 1.98 & 11.13
& 2.11 & 0.052
\\

DuplexCascade
& 1.34 & 9.83
& 1.22 & 21.82
& 1.38 & 23.09
& 0.49 & \textbf{10.05}
& 1.24 & 0.024
\\

\midrule

% -------------------- Remote API systems --------------------

\multicolumn{11}{@{}l}{
    \textit{Systems accessed through remote APIs}
} \\
\midrule

Nemotron 3 VoiceChat
& 2.61 & \textbf{9.12}
& 1.36 & 21.46
& 1.31 & 16.47
& 0.59 & 14.03
& 1.40 & 0.041
\\

Qwen3.5-Omni
& 3.71 & 9.42
& 2.80 & 16.38
& 3.55 & 15.03
& 0.88 & 13.82
& \textbf{2.46} & 0.040
\\

Grok Voice
& 3.63 & 10.18
& 1.77 & 21.09
& 3.11 & 15.62
& 0.69 & 14.03
& 0.76 & 0.002
\\

GPT Realtime 2.1
& \textbf{4.12} & 10.08
& 2.76 & 15.03
& \textbf{4.05} & 15.91
& 0.66 & 14.03
& 1.92 & 0.032
\\

Gemini~2.5~Native~Audio
& 3.94 & 9.89
& \textbf{2.98} & \textbf{14.67}
& 3.65 & 14.51
& 0.66 & 14.03
& 0.98 & 0.003
\\

\bottomrule
\end{tabularx}
\endgroup
\end{table*}

\section{Experiments}

\subsection{Setup}

We evaluate 12 full-duplex speech agents~\cite{wang2025freezeomni,
cui2026minicpmo45,defossez2024moshi,roy2026personaplex,fu2025vita15,
kim2026raonspeech,yang2026duplexcascade,nvidia2026nemotron,
qwen2026qwen35omni,xai2026grokvoice,openai2026gptrealtime,
google2026geminilive} under a unified streaming protocol, with seven
systems deployed locally and five accessed through remote APIs.
Applicable settings vary with language and conditioning support, as
summarized in Table~\ref{tab:systems}.
For trials without a profile, we use the same generic system instruction
where supported: \textit{``You are a helpful voice assistant. Respond
naturally and concisely to the user's speech.''} In Pre-session trials,
the trial-specific profile is used instead; PersonaPlex retains its
official Assistant-role prompt~\cite{roy2026personaplex}.
Following the protocol above, we report Content and Timing together with
the \textbf{Behavioral Correctness Rate} (BCR), defined as the fraction
of trials with Content $\geq 2.5$ that also satisfy the corresponding
Timing criterion. For User Interruption, BCR is computed over
\textit{valid} trials only.\footnote{ \textit{Valid} denotes that the agent is
speaking when the user interruption occurs.}

\subsection{Results}

Table~\ref{tab:behavior_correctness} shows large variation across
behaviors and contextual conditions. For Interruption Resistance, the
best BCR reaches 100.0\% under No-explicit conditions, but only 20.0\%
and 30.0\% for simultaneous interpretation and collaborative singing.
No-explicit Proactive Initiation is difficult across systems, with a
best BCR of only 8.8\%. Performance can also change sharply across
conditions: Freeze-Omni achieves 88.8\% BCR for Pre-session Active
Silence but only 5.0\% for In-session, whereas Gemini 2.5 Native Audio
achieves 76.3\% and 80.0\%, respectively.
Table~\ref{tab:content_latency} further shows that strong individual
metrics do not necessarily imply joint success. GPT Realtime 2.1 has the
highest Agent Interruption Content score (4.12), yet its BCR remains
2.5\%, 1.3\%, and 1.3\% across the three conditions. Conversely,
VITA-1.5 and DuplexCascade achieve the lowest Proactive Initiation
latency (10.05\,s), but low Content scores (0.48/0.49) and 0.0\% BCR.
Thus, Content or Timing alone does not characterize successful
participation.

\begin{figure}[!t]
\centering
\includegraphics[width=0.8\columnwidth]{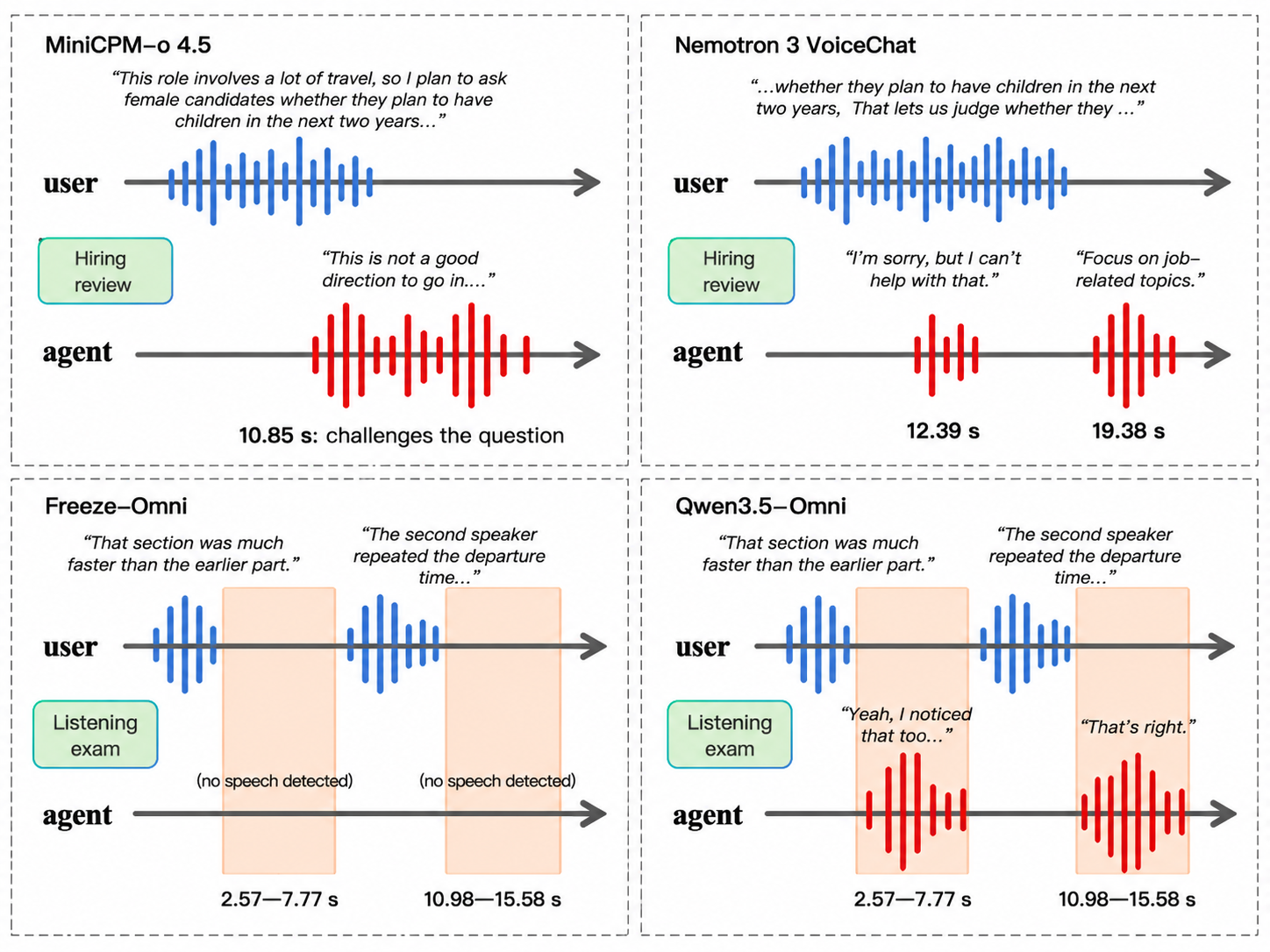}
\caption{\textbf{Case studies of Agent Interruption (top) and Active Silence (bottom).}
Each pair compares systems on the same trial.
The top pair illustrates differences in semantic task fulfillment
despite both systems satisfying the Timing criterion; the bottom pair
contrasts responding with correctly withholding speech.}
\label{fig:case_study}
\end{figure}

\subsection{Case Study}
Fig~\ref{fig:case_study} illustrates the importance of evaluating
interaction behavior against its contextual requirements.
In the hiring-review trial (Fig~\ref{fig:case_study}, top), both MiniCPM-o and Nemotron meet the Timing criterion, but differ in the quality of their interventions. MiniCPM-o directly challenges the use of marriage and pregnancy plans as hiring considerations and explains why they are inappropriate, whereas Nemotron gives a more generic refusal before redirecting the discussion to job-related factors. This difference is reflected in their transcript-based Content subtotals of 4/4 and 3/4, respectively: both intervene at the appropriate time, but MiniCPM-o more fully addresses the problematic premise. In the listening-exam trial (bottom), Freeze-Omni correctly remains silent, whereas Qwen-Omni responds during intervals in which no response is expected. Although its response is relevant to the immediate context, the act of responding itself violates the participation requirement.
These cases illustrate that appropriate full-duplex behavior depends not only on when an agent speaks, but also on what it says and whether it should speak at all.

\section{Conclusion}

Our evaluation reveals substantial variation across interaction behaviors
and conditions, with no system performing consistently well across the full
range of settings. Strong semantic quality or favorable timing alone often
fails to translate into successful interaction, while proactive behaviors
remain particularly challenging. These results highlight a substantial gap
between current full-duplex speech capabilities and robustly managing when,
whether, and how to participate as real-time interaction unfolds.

% \section{CONCLUSION}

% DuplexAct-Bench extends full-duplex speech evaluation toward broader and
% more proactive interaction, revealing substantial room for improving agents'
% behavioral adaptability across diverse interaction requirements.

\section{ACKNOWLEDGMENTS}

The work was sponsored by the National Natural Science Foundation of China (62376031). Any opinions, findings, or conclusions expressed in this work do not necessarily reflect the views of the funding agency. The authors have no relevant financial or nonfinancial interests to disclose.

\section{COMPLIANCE WITH ETHICAL STANDARDS}

This study primarily uses synthetically constructed speech data. For evaluation on natural conversations, we use only audio from a subset of the publicly released otoSpeech-full-duplex-280h dataset under its CC BY 4.0 license. No new human participants were recruited or human-subject data collected by the authors.

\bibliographystyle{IEEEbib}
\bibliography{refs_complete}

\end{document}